\documentclass[a4paper]{article}

\usepackage{INTERSPEECH2018}

\usepackage{graphicx}
\usepackage{amssymb,amsmath,bm}
\usepackage{textcomp}
\usepackage[inline]{enumitem}
\usepackage{graphics}
\usepackage{multirow}
\usepackage{hyperref}

\usepackage{algpseudocode}
\usepackage{algorithm}

\def\vec#1{\ensuremath{\bm{{#1}}}}

\def\x{\bm{x}}
\def\X{\bm{X}}
\def\z{\bm{z}}
\def\Z{\bm{Z}}
\def\bmu{\bm{\mu}}

\usepackage{xcolor}

\eightpt

\title{Scalable Factorized Hierarchical Variational Autoencoder Training}

\name{Wei-Ning Hsu, James Glass}
\address{Computer Science and Artificial Intelligence Laboratory \\
         Massachusetts Institute of Technology\\
         Cambridge, MA 02139, USA}
\email{\{wnhsu,glass\}@mit.edu}

\begin{document}

  \maketitle
  
  \begin{abstract}
  Deep generative models have achieved great success in unsupervised learning with the ability to capture complex nonlinear relationships between latent generating factors and observations. 
  Among them, a factorized hierarchical variational autoencoder (FHVAE) is a variational inference-based model that formulates a hierarchical generative process for sequential data.
  Specifically, an FHVAE model can learn disentangled and interpretable representations, which have been proven useful for numerous speech applications, such as speaker verification, robust speech recognition, and voice conversion.
  However, as we will elaborate in this paper, the training algorithm proposed in the original paper is not scalable to datasets of thousands of hours, which makes this model less applicable on a larger scale.
  After identifying limitations in terms of runtime, memory, and hyperparameter optimization, we propose a hierarchical sampling training algorithm to address all three issues.
  Our proposed method is evaluated comprehensively on a wide variety of datasets, ranging from 3 to 1,000 hours and involving different types of generating factors, such as recording conditions and noise types.
  In addition, we also present a new visualization method for qualitatively evaluating the performance with respect to the interpretability and disentanglement.
  Models trained with our proposed algorithm demonstrate the desired characteristics on all the datasets.
  \end{abstract}

  \section{Introduction}
  Unsupervised learning can leverage large amounts of unlabeled data to discover latent generating factors that often lie on a lower dimensional manifold compared to the raw data.
  A learned latent representation from speech can be useful for many downstream applications, such speaker verification~\cite{dehak2011front}, automatic speech recognition~\cite{tan2016learning}, and linguistic unit discovery~\cite{drexler2016deep,hsu2017learning}.
  A factorized hierarchical variational autoencoder (FHVAE)~\cite{hsu2017unsupervised} is a variational inference-based deep generative model that learns interpretable and disentangled latent representation from sequential data without supervision by modeling a hierarchical generative process.
  In particular, it has been demonstrated that an FHVAE trained on speech data learns to encode sequence-level generating factors, such as speaker and channel condition, into one set of latent variables, while encoding segment-level generating factors, such as phonetic content, into another set of latent variables.
  The ability to disentangle latent factors has been beneficial to a wide range of tasks, including domain adaptation~\cite{hsu2018extracting}, conditional data augmentation~\cite{hsu2017unsuperviseddomain}, and voice conversion~\cite{hsu2017unsupervised}.
  
  However, the original FHVAE training algorithm proposed in~\cite{hsu2017unsupervised} does not scale to datasets of over hundreds of thousands of utterances, making it less applicable to real world settings, where an unlabeled dataset of such size is common.
  This limitation is mainly due to the following issues:
  \begin{enumerate*}[label=(\arabic*)]
    \item the inference model of the sequence-level latent variable, and
    \item the design of the discriminative objective.
  \end{enumerate*}
  To be more specific, the original training algorithm reduces the complexity of inferring sequence-level latent variables by maintaining a cache, whose number of entries equals the number of training sequences.
  In addition, the discriminative objective, which encourages disentanglement, requires computing a partition function that sums over the entries in that cache.
  The two facts combined lead to significant scalability issues.
  
  In this paper, we propose a hierarchical sampling algorithm to address these issues.
  In addition, a new method for qualitatively evaluating disentanglement performance based on a t-Distribution Stochastic Neighbor Embedding~\cite{van2014accelerating} is also presented.
  The proposed training algorithm is evaluated on a wide variety of datasets, ranging from 3 to 1,000 hours and involving many different types of generating factors, such as recording conditions and noise types.
  Experimental results verify that the proposed algorithm is effective on all sizes of datasets and achieves desirable disentanglement performance.
  The code is available on-line.\footnote{https://github.com/wnhsu/ScalableFHVAE}
  
  \section{Limitations of Original FHVAE Training}
  In this section, we briefly review the factorized hierarchical variational autoencoder (FHVAE) model and discuss the scalability issues of the original training objective.
  
  \subsection{Factorized Hierarchical Variational Autoencoders}
  An FHVAE~\cite{hsu2017unsupervised} is a variant of a VAE that models a generative process of sequential data with a hierarchical graphical model.
  Let $\X = \{ \x^{(n)} \}_{n=1}^N$ be a sequence of $N$ segments.
  An FHVAE assumes that the generation of a segment $\x$ is conditioned on a pair of latent variables, $\z_1$ and $\z_2$, referred to as the \textit{latent segment variable} and the \textit{latent sequence variable} respectively.
  While $\z_1$ is generated from a global prior, similar to those latent variables in a vanilla VAE, $\z_2$ is generated from a sequence-dependent prior that is conditioned on a sequence-level latent variable, $\bmu_2$, named the \textit{s-vector}.
  The joint probability of a sequence is formulated as: $p(\bmu_2) \prod_{n=1}^N p(\x^{(n)} | \z_1^{(n)}, \z_2^{(n)}) p(\z_1^{(n)}) p(\z_2^{(n)} | \bmu_2)$.
  With this formulation, an FHVAE learns to encode generating factors that are consistent within segments drawn from the same sequence into $\z_2$.
  In contrast, $\z_1$ captures the residual generating factors that changes between segments.
  
  Since computing the true posteriors of $\Z_1 = \{\z_1^{(n)}\}_{n=1}^N$, $\Z_2 = \{\z_2^{(n)}\}_{n=1}^N$, and $\bmu_2$ are intractable, an FHVAE introduces an inference model, $q(\Z_1, \Z_2, \bmu_2 | \X)$, and factorizes it as: $q(\bmu_2 | \X) \prod_{n=1}^N q(\z_1^{(n)} | \x^{(n)}, \z_2^{(n)}) q(\z_2^{(n)} | \x^{(n)})$.
  We summarize in Table~\ref{tab:fhvae} the family of distributions an FHVAE adopts for the generative model and the inference model.
  All the functions, $f_{\mu_{\x}}$, $f_{\sigma_{\x}^2}$, $g_{\mu_{\z_1}}$, $g_{\sigma_{\z_1}^2}$, $g_{\mu_{\z_2}}$, and $g_{\sigma_{\z_2}^2}$, are neural networks that parameterize mean and variance of Gaussian distributions.
  
  \begin{table}[tbh]
  	\centering
    \caption{Family of distributions adopted for FHVAE generative and inference models.}
    \begin{tabular}{l|l}
    \hline\hline
    \multicolumn{2}{c}{\textit{generative model}} \\
    \hline
    $p(\bmu_2)$ & $\mathcal{N}(\bm{0}, \bm{I})$ \\
    $p(\z_1)$ & $\mathcal{N}(\bm{0}, \bm{I})$ \\
    $p(\z_2 | \bmu_2)$ & $\mathcal{N}(\bmu_2, \sigma_{\z_2}^2\bm{I})$ \\
    $p(\x | \z_1, \z_2) $ & $\mathcal{N}(f_{\mu_{\x}}(\z_1, \z_2), diag(f_{\sigma_{\x}^2}(\z_1, \z_2)))$ \\
    \hline\hline
    \multicolumn{2}{c}{\textit{inference model}} \\
    \hline
    $q(\bmu_2 | \X)$ & $\mathcal{N}(\sum_{n=1}^N g_{\mu_{\z_2}}(\x^{(n)})/(N + \sigma_{\z_2}^2), \bm{I})$\\
    $q(\z_1 | \x, \z_2)$ & $\mathcal{N}(g_{\mu_{\z_1}}(\x, \z_2), diag(g_{\sigma_{\z_1}^2}(\x, \z_2)))$ \\
    $q(\z_2 | \x)$ & $\mathcal{N}(g_{\mu_{\z_2}}(\x), diag(g_{\sigma_{\z_2}^2}(\x)))$ \\
    \hline\hline
    \end{tabular}
    \label{tab:fhvae} 
  \end{table}
  
  \subsection{Original FHVAE Training}
  In the variational inference framework, since the marginal likelihood of observed data is intractable, we optimize the variational lower bound, $\mathcal{L}(p, q; \X)$, instead.
  We can derive a sequence variational lower bound of $\X$ based on Table~\ref{tab:fhvae}.
  However, this lower bound can only be optimized at the sequence level, because inferring of $\bmu_2$ depends on an entire sequence, and would become infeasible if $\X$ is extremely long.
  
  In~\cite{hsu2017unsupervised} the authors proposed replacing the maximum a posterior (MAP) estimation of $\bmu_2$'s posterior mean for training sequences with a cache $h_{\mu_{\bmu_2}}(i)$, where $i$ indexes training sequences. 
  In other words, the inference model for $\bmu_2$ becomes $q(\bmu_2 | \X^{(i)}) = \mathcal{N}(h_{\mu_{\bmu_2}}(i), \bm{I})$.
  Therefore, the lower bound can be re-written as: 
  \begin{align}
    \mathcal{L}&(p, q; \X^{(i)}) = \sum_{n=1}^{N^{(i)}} \mathcal{L}(p, q; \x^{(i, n)} | h_{\mu_{\bmu_2}}(i)) + \log p(h_{\mu_{\bmu_2}}(i)) \\
    \mathcal{L}&(p, q; \x^{(i, n)} | h_{\mu_{\bmu_2}}(i)) = \mathbb{E}_{q(\z_1, \z_2 | \x^{(i,n)})} [ \log p(\x^{(i,n)} | \z_1, \z_2)] \nonumber \\
    & - \mathbb{E}_{q(\z_2 | \x^{(i,n)})} [ D_{KL} (q(\z_1 | \x^{(i,n)}, \z_2) || p(\z_1)) ]  \nonumber \\
    & - D_{KL} (q(\z_2 | \x^{(i,n)}) || p(\z_2 | h_{\mu_{\bmu_2}}(i))). 
  \end{align}
  We can now sample a batch at the segment level to optimize the following variational lower bound:
  \begin{equation}
  \mathcal{L}(p, q; \x^{(i, n)}) = \mathcal{L}(p, q; \x^{(i, n)} | h_{\mu_{\bmu_2}}(i)) + \dfrac{1}{{N^{(i)}}} \log p(h_{\mu_{\bmu_2}}(i)). 
  \end{equation}
  
  Furthermore, to obtain meaningful disentanglement between $\z_1$ and $\z_2$, it is not desirable to have constant $\bmu_2$ for all sequences; otherwise, for each segment, swapping $z_1$, $z_2$ would lead to the same objective.
  To avoid such condition, the following objective is added to encourage $\bmu_2$ to be discriminative between sequences:
  \begin{equation}
    \log p(i | \bar{\z}_2^{(i,n)} ) := 
    \log \dfrac{p( \bar{\z}_2^{(i,n)}  | \bar{\bmu}_2^{(i)} ) }{\sum_{j=1}^M p( \bar{\z}_2^{(i,n)}  | \bar{\bmu}_2^{(j)} )}, 
  \end{equation}
  where $M$ is the total number of training sequences, $\bar{\z}_2^{(i,n)} = g_{\mu_{\z_2}}(\x^{(i,n)})$ denotes the posterior mean of $\z_2$, and $\bar{\bmu}_2^{(i)} = h_{\mu_{\bmu_2}}(i)$ denotes the posterior mean of $\bmu_2$.
  This additional discriminative objective encourages $\z_2$ from the $i$-th sequence to be not only close to $\bmu_2$ of the $i$-th sequence, 
  but also far from $\bmu_2$ of other sequences.
  Combining this discriminative objective and the segment variational lower bound with a weighting parameter, $\alpha$, the objective function that an FHVAE maximizes then becomes:
  \begin{equation}
    \mathcal{L}^{dis}(p, q; \x^{(i,n)}) = 
    \mathcal{L}(p, q; \x^{(i, n)}) + \alpha \log p(i | \bar{\z}_2^{(i,n)} ), \label{eq:dis}
  \end{equation}
  referred to as the \textit{discriminative segment variational lower bound}.
  
  \subsection{Scalability Issues}
  The original FHVAE training addressed the scalability issue with respect to sequence length by decomposing a sequence variational lower bound into sum of segmental variational lower bounds over segments.
  However, here we will show that this training objective is not scalable with respect to the number of training sequences.
  
  First of all, the original FHVAE training maintains an $M$-entry cache, $h_{\mu_{\bmu_2}}(\cdot)$, that stores the posterior mean of $\bmu_2$ for each training sequence.
  The size of this cache scales linearly in the number of training sequences.
  Suppose $\z_2$ are 32-dimensional 32-bit floating point vectors as in~\cite{hsu2017unsupervised}.
  If the number of training sequences is on the order of $10^8$, the cache size would grow to about $10$ GB and exhaust the memory of a typical commercial GPU ($8$ GB).
  Even worse, when computing the gradient given a batch of training segments, we need to maintain a tensor of size $(bs, |\theta|)$, where $bs$ is the segment batch size, and $|\theta|$ is the total number of trainable parameters involved in the computation of the objective function.
  Since the computation of the discriminative objective involves the entire cache, $h_{\mu_{\bmu_2}}(\cdot)$, the gradient tensor is of size at least $bs$ times larger than the cache.
  With a batch size of 256, a dataset with $10^5$ sequences can exhaust the GPU memory during training.

  Second, the denominator of the discriminative objective, $\sum_{i=1}^M p( \bar{\z}_2^{(i,n)}  | \bar{\bmu}_2^{(j)} )$, marginalizes over a function of posterior mean of $\bmu_2$ of all training sequences, which increases the computation time proportionally to the number of sequences for each training step.
  
\begin{figure}[t]
	\centerline{\includegraphics[width=.83\linewidth]{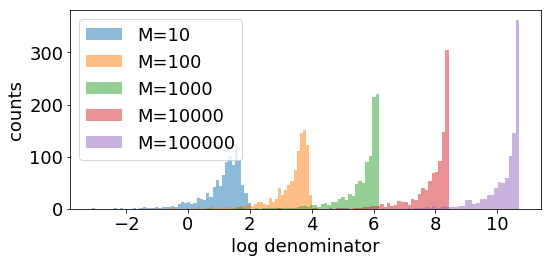}}
	\caption{Histogram of $\log \sum_{i=1}^M p( \bar{\z}_2^{(i,n)}  | \bar{\bmu}_2^{(j)} )$ with respect to different $M\in \{ 10^1, 10^2, 10^3, 10^4, 10^5\}$. Distributions shift by roughly a constant when $M$ increases by 10 times, implying the denominator scales proportionally to $M$.}
	\label{fig:dis_den}
  \end{figure}

  Third, from the hyperparameter optimization point of view, the distribution of the denominator, $\sum_{i=1}^M p( \bar{\z}_2^{(i,n)}  | \bar{\bmu}_2^{(j)} )$, also changes with respect to the number of training sequences.
  Specifically, the expected value of this terms scales roughly linearly in $M$, as shown in Figure~\ref{fig:dis_den}.
  Such behavior is not desirable, because the $\alpha$ parameter that balances the variational lower bound and the discriminative objective would need to be adjusted according to $M$.

  \section{Training with Hierarchical Sampling}
  In order to utilize the discriminative objective, while eliminating the memory, computation, and optimization issue induced by a large training set, we need to control the size of the cache as well as the denominator summation in the discriminative objective.
  Both of these can be achieved jointly with a hierarchical sampling algorithm.
  
  Given a dataset of $M$ training sequences, \textbf{we maintain a cache of only $K$ entries, where $K$ is a dataset independent hyperparameter, named the \textit{sequence batch size}}.
  We optimize an FHVAE model by repeating the following procedure until the convergence criterion is met:
  \begin{enumerate*}[label=(\arabic*)]
  	\item Sample a batch of $K$ sequences from the entire training set.
	\item Reset each entry of the cache, $h_{\mu_{\bmu_2}}(k)$, with the MAP estimation, $\sum_{n=1}^{\tilde{N}^{(k)}} g_{\mu_{\z_2}}(\tilde{\x}^{(k, n)})/(\tilde{N}^{(k)} + \sigma_{\z_2}^2)$, where $\tilde{N}^{(k)}$ is the number of segments in the $k$-th sampled sequence, and $\tilde{\x}^{(k, n)}$ is the $n$-th segment of the $k$-th sampled sequence, for $k = 1, \dots, K$. 
    \item Sample $B_{seg}$ batches of segments sequentially from the $K$ sampled sequences. Each batch of segments is used to estimate the discriminative segmental variational lower bound for optimizing the parameters of $f_*$, $g_*$, and $h_{\mu_{\bmu_2}}$ as before.
  	The only difference is that \textit{the denominator of the discriminative object now sums over the $K$ sampled training sequences}, instead of the entire set of $M$ training sequences.
  \end{enumerate*}  
  We list the pseudo code in Algorithm~\ref{alg:hs}.
  
  \begin{algorithm}[t]
  \caption{Training with Hierarchical Sampling}\label{alg:hs}
  \textbf{Input:} $\{ \X^{(i)} \}_{i=1}^M$: training set; $K$: sequence batch size; $bs$: segment batch size; $B_{seg}$: number of segment batches; $f_*$/$g_*$: decoders/encoders; $h_{\mu_{\bmu_2}}$: cache of $K$ entries; Optim: gradient descent-based optimizer
  \begin{algorithmic}[1]
  \While {not converged}
  	\State sample a batch of $K$ training sequences, $\{ \tilde{\X}^{(k)} \}_{k=1}^K$
  	\For {$k = 1 \dots K$}
      \State $h_{\mu_{\bmu_2}}(k) \gets \sum_{n=1}^{N^{(k)}} g_{\mu_{\z_2}}(\tilde{\x}^{(k, n)})/(N + \sigma_{\z_2}^2)$
  	\EndFor
    \For {$1 \dots B_{seg}$}
      \State sample segments $\{ \tilde{\x}^{(k_b, n_b)} \}_{b=1}^{bs}$ from $\{ \tilde{\X}^{(k)} \}_{k=1}^K$
      \State $l_{dis}(b) \gets - \log \dfrac{p( g_{\mu_{\z_2}}(\tilde{\x}^{(k_b,n_b)})  | h_{\mu_{\bmu_2}}(k_b) ) }{\sum_{k=1}^K p( g_{\mu_{\z_2}}(\tilde{\x}^{(k_b,n_b)})  | h_{\mu_{\bmu_2}}(k) )} $
      \State $l_{gen}(b) \gets - \mathcal{L}(p, q; \tilde{\x}^{(k_b,n_b)})$
	  \State $loss \gets \sum_{b=1}^{bs} (l_{gen}(b) + \alpha \cdot l_{dis}(b) ) / {bs}$
  	  \State $f_*, g_*, h_{\mu_{\bmu_2}} \gets \text{Optim}(loss, \{ f_*, g_*, h_{\mu_{\bmu_2}} \})$
    \EndFor
  \EndWhile
  \State \textbf{return} $f_*, g_*$
  \end{algorithmic}
  \end{algorithm}
  
  We refer to the proposed algorithm as a hierarchical sampling algorithm, because we first sample at the sequence level, and then at the segment level.
  The size of the cache is then controlled by the sequence batch size $K$, instead of the number of training sequences $M$.
  The algorithm can also be viewed as iteratively sampling a sub-dataset of $K$ sequences and running the original training algorithm on it.
  Compared with the proposed algorithm, the original training algorithm can be regarded as a ``flat'' sampling algorithm, where we sample segments from the entire pool, so it is therefore necessary to maintain a cache of $M$ entries.
  The proposed algorithm introduces an overhead associated with reseting the cache whenever a new batch of sequences is sampled.
  However, this cost can be amortized by increasing the number of segment batches $B_{seg}$, for each batch of sequences.
    
  \section{Experimental Setup}
  We evaluate our training algorithm on a wide variety of datasets, ranging from 3 to 1,000 hours, including both clean and noisy, close-talking and distant speech.
  In this section, we describe the datasets, and introduce FHAVE models and their training configurations.
  
  \subsection{Datasets}
  Four different corpora are used for our experiments, which are TIMIT~\cite{garofolo1993darpa}, Aurora-4~\cite{pearce2002aurora}, AMI~\cite{carletta2007unleashing}, and LibriSpeech~\cite{panayotov2015librispeech}.
  TIMIT contains broadband 16kHz recordings of phonetically-balanced read speech.
  A total of 3,696 utterances (3 hours) are presented in the training partition based on the Kaldi~\cite{povey2011kaldi} recipe, where \texttt{sa} utterances are excluded.
  In addition, manually labeled time-aligned phonetic transcripts are available, which we use to study the disentanglement performance between phonetic and speaker information achieved by FHVAE models.
  
  Aurora-4 is another broadband corpus designed for noisy speech recognition tasks, based on the Wall Street Journal corpus (WSJ0)~\cite{garofalo2007csr}.
  Six different types of noise are artificially mixed with clean read speech of two different microphone types, amounting to a total of 4,620 utterances (10 hours) for the development set.
  Following the training setup in~\cite{hsu2017unsupervised}, we train our FHVAE models on the development set, because the noise and channel information on the training set is not available.
  
  The AMI corpus consists of 100 hours of meeting recordings, recorded in three different meeting rooms with different acoustic properties, and with multiple attendants. 
  Multiple microphones are used for each session, including individual headset microphones (IHM), and far-field microphone arrays.
  IHM and single distant microphone (SDM) recordings from the training set are mixed to form a training set for the FHVAE models, including over 200,000 utterances according to the segmentation provided in the corpus.
  
  The largest corpus we evaluate on is the LibriSpeech corpus, which contains 1,000 hours of read speech sampled at 16kHz.
  This corpus is based on the LibriVox's project, where world-wide volunteers record public domain texts to create free public domain audio books.

  \begin{figure}[t]
	\centerline{\includegraphics[width=\linewidth]{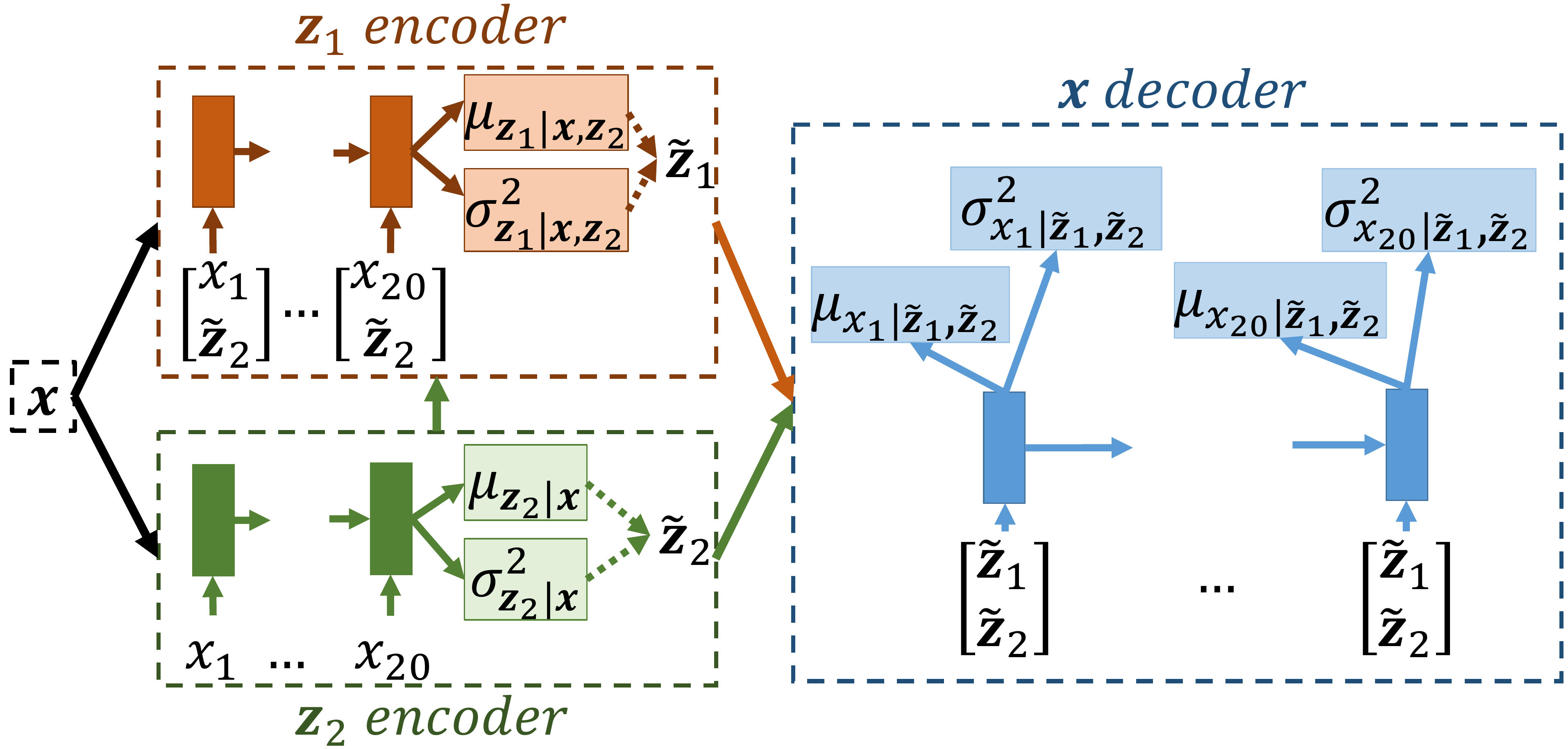}}
    \caption{The proposed FHVAE architecture consists of two encoders (orange and green) and one decoder (blue). $\x = [x_1, \cdots, x_{20}]$ is a segment of 20 frames. Dotted lines in the encoders denote sampling from parametric distributions.}
	\label{fig:arch}
  \end{figure}

  \subsection{Training and Model Configurations}
  Speech segments of 20 frames, represented with 80-dimensional log Mel-scale filter bank coefficients (FBank), are used as inputs to FHVAE models.
  We denote each segment with $\x = [x_1, \cdots, x_{20}]$. 
  The variance of  $\z_2$'s prior is set to $\sigma^2_{\z_2}=0.25$, and the dimension of $\z_1$ and $\z_2$ are both 32.
  Figure~\ref{fig:arch} illustrates the detailed encoder/decoder architectures of the proposed FHVAE model.
  The conditional mean and variance predictor for each variable (i.e., $\z_1$, $\z_2$, and $\x$) shares a common stacked LSTM pre-network, followed by two different single-layer affine transform networks, $\mu_*$ and $\sigma^2_*$, predicting the conditional mean and variance respectively.
  Specifically, a stacked LSTM with 2 layers and 256 memory cells are used for all three pre-networks, illustrated in Figure~\ref{fig:arch} with blocks filled with dark colors.
  Affine transform networks of $\z_1$ and $\z_2$ encoders take as input the output from the last time step of both layers, which sums to 512 dimension.
  As for the $\x$ decoder, the affine transform network takes as input the LSTM output of the last layer from each time step $t$, and predicts the probability distribution of the corresponding frame $p(x_t | \bm{z}_1, \bm{z}_2)$.
  The same sampled $\tilde{\z}_1$ and $\tilde{\z}_2$ from the posterior distributions are concatenated and used as input for the LSTM decoder at each step.
  Sampling is done by introducing auxiliary input variables for the reparameterization trick~\cite{kingma2013auto}, in order to keep the entire network differentiable with respect to the objective.

  FHVAE models are trained to optimize the discriminative segment variational lower bound with $\alpha=10$.
  We set sequence batch size $K=2000$ for TIMIT and Aurora-4, and $K=5000$ for the others.
  Adam~\cite{kingma2014adam} with $\beta_1 = 0.95$ and $\beta_2 = 0.999$ is used to optimize all models.
  Tensorflow~\cite{abadi2016tensorflow} is used for implementation.
  Training is done for 500,000 steps, terminating early if the segmental variational lower bound on a held-out validation set is not improved for 50,000 steps.
  
  \begin{figure*}[t]
	\centerline{\includegraphics[width=\linewidth]{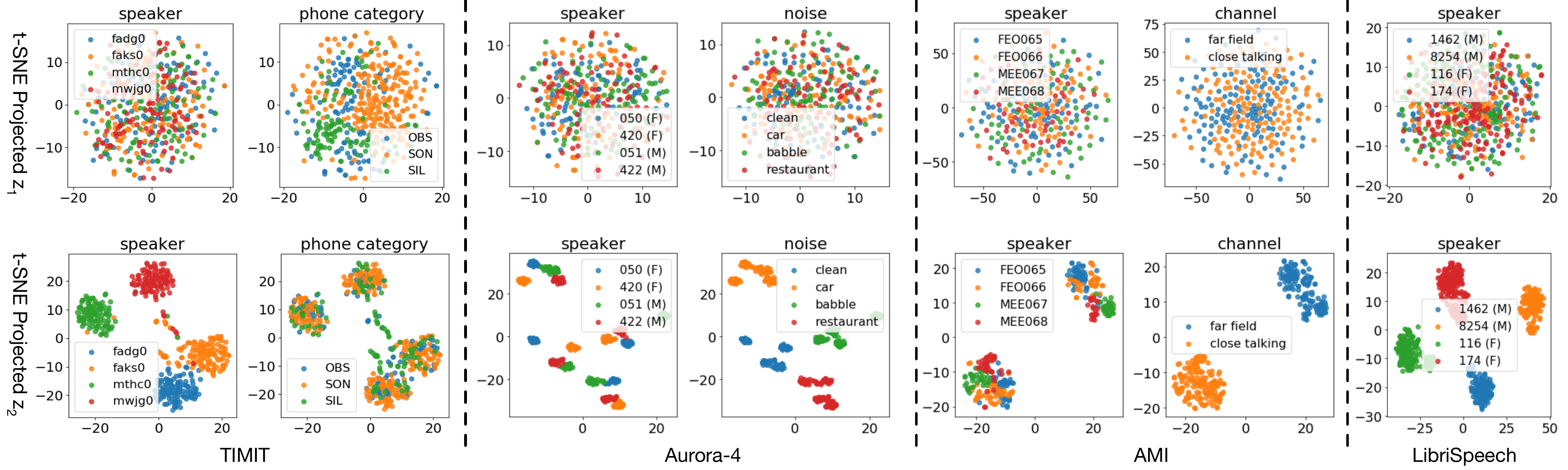}}
    \caption{Scatter plots of t-SNE projected $\z_1$ and $\z_2$ with models trained on TIMIT/Aurora-4/AMI/LibriSpeech. Each point represents one segment. Different colors are used to code segments of different labels with respect to the generating factor shown at the title of each plot.}
	\label{fig:tsne}
  \end{figure*}
  
  \section{Results and Discussions}
  \subsection{Time and Memory Complexity}
  One feature of our proposed training algorithm is to control memory complexity.
  We found that a training set with over 100,000 sequences would exhaust a single 8GB GPU memory when using the original training algorithm.
  Hence, it was not feasible for the AMI and the LibriSpeech corpus, while the proposed algorithm does not suffer from the same problem.
  Another feature of hierarchical sampling is to control the time complexity of computing the discriminative loss.
  To study how sequence batch size affects the optimization step (line 8 in Algorithm~\ref{alg:hs}), we evaluate the processing time of that step by varying $K$ from 20 to 20,000 and show the results in Table~\ref{tab:time}.
  
  We can observe that when $K \le 2000$, the time complexity of computing the discriminative loss is fractional compared to computing the variational lower bound.
  However, when $K > 2000$, the increased computation time grows proportional to the sequence batch size, so that computation of the discriminative loss starts to dominate the time complexity. 
  In practice, given a new encoder/decoder architecture, we can investigate the computation overhead resulting from the discriminative loss using such a method, and it is possible to determine some $K$ that introduces negligible overhead for optimization.
  \begin{table}[h]
    \centering
    \caption{Processing time of the optimization step with different sequence batch size $K$.}
    \resizebox{\linewidth}{!}{
    \begin{tabular}{c|cccccccccc}
    $K$ & 10 & 100 & 1000 & 2000 & 5000 & 10000 & 20000 \\
    \hline
    Time (ms) & 84 & 84 & 86 & 87 & 103 & 147 & 230
    \end{tabular}
    }
  	\label{tab:time}
  \end{table}
  
  \subsection{Evaluating Disentanglement Performance}
  To examine whether an FHVAE is successfully trained, we need to inspect its performance at disentangling sequence-level generating factors (e.g. speaker identity, noise condition, and channel condition) from segment-level generating factors (e.g. phonetic content) in the latent space. 
  For quantitative evaluation, we reproduce the speaker verification experiments in~\cite{hsu2017unsupervised}.
  The FHVAE model trained with hierarchical sampling achieves 1.64\% equal error rate on TIMIT, matching the performance of the original training algorithm (1.34\%).
  In the following sections, we proceed with two qualitative evaluation methods.
  
  \subsubsection{t-SNE Visualization of Latent Variables}
  We start with selecting a batch of labeled segments $( \x, y )$, where $y$ denotes the values of the associated generating factors, for example $y$ = (phone-id, speaker-id).
  We then infer $\z_1$ and $\z_2$ of these segments, and project them separately to a two-dimensional space using t-Distributed Stochastic Neighbor Embedding (t-SNE)~\cite{van2014accelerating}.
  Each generating factor is used to color-code both projected $\z_1$ and $\z_2$.
  Successful disentanglement would result in segments of the same sequence-level generating factors forming clusters in the projected $\z_2$ space but not in the projected $\z_1$ space, and vice versa.
  
  For all four datasets, speaker label, a sequence-level generating factor, is available for each segment.
  Since time-aligned phonetic transcripts are available for TIMIT, it is also possible to derive phone labels, which is a segment-level generating factor.
  Following~\cite{halberstadt1999heterogeneous}, we further reduce the 61 phonemes to three phonetic subsets: sonorant (SON), obstruent (OBS), and silence (SIL) for better color-coding. 
  In addition, noise types can be obtained for Aurora-4, and microphone types can be obtained for AMI, which are both sequence-level generating factors.
  
  Results of t-SNE projections for models trained on each dataset are shown in Figure~\ref{fig:tsne}, where each point represents one segment.
  It can be observed that in each of the projected $\z_2$ spaces, segments of the same sequence-level generating factors (speaker/noise/channel) always form clusters.
  When segments are generated conditioned on multiple sequence-level generating factors, as in Aurora-4 and AMI, the segments actually cluster hierarchically. 
 In contrast, the distribution of projected $\z_1$'s does not vary between different values of these generating factors, which implies that $\z_1$ does not contain much information about them.
  Opposite phenomenon can be observed from the phone category-coded plots, where segments belonging to the same phonetic subset cluster in the projected $\z_1$ space, but not in the projected $\z_2$ space. 
  These results suggest that FHVAEs trained with hierarchical sampling can achieve desirable disentanglement for these conditions.

  \subsubsection{Reconstructing Re-combined Latent Variables}
  \begin{figure}[t]
	\centerline{\includegraphics[width=\linewidth]{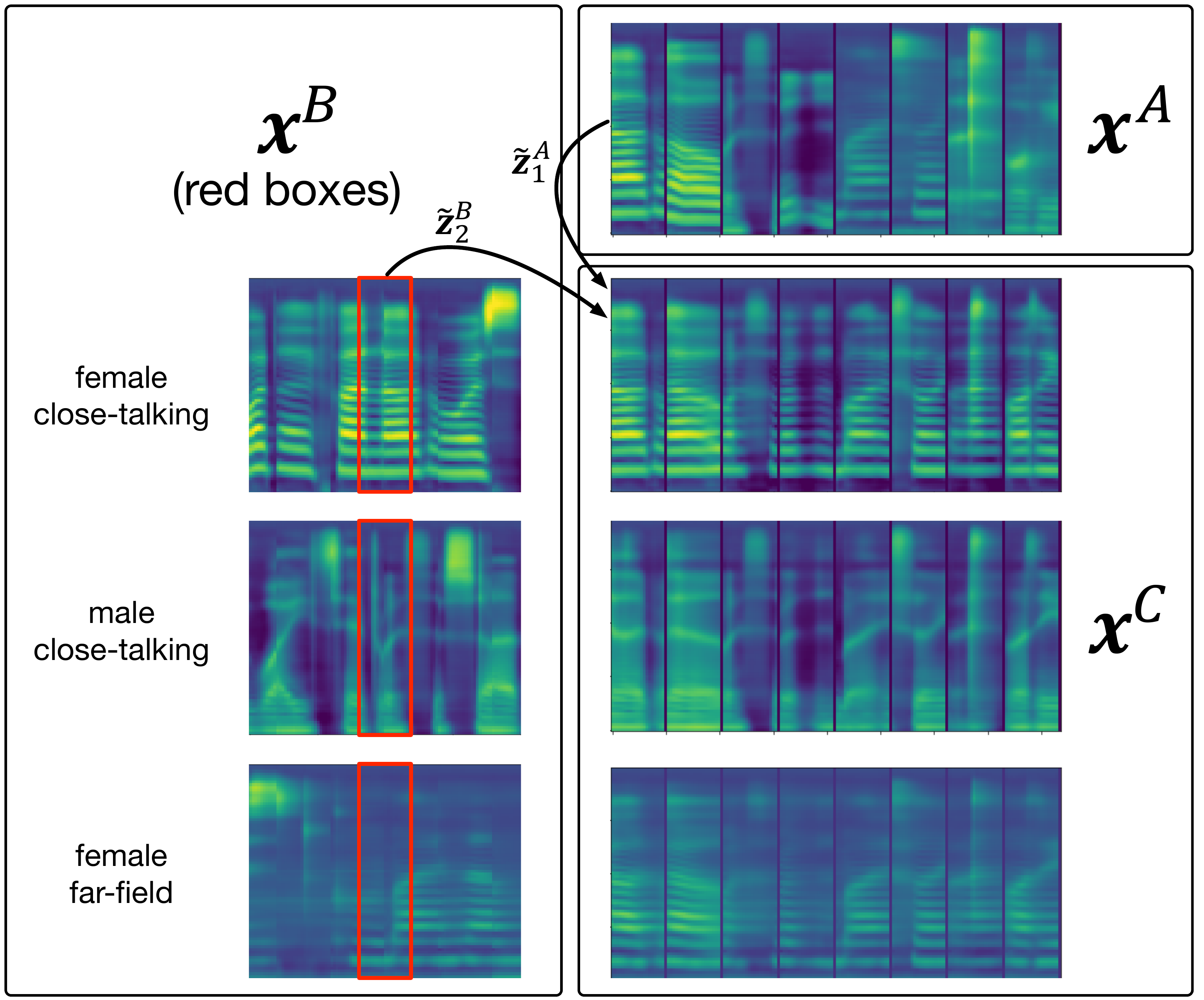}}
    \caption{Results of decoding re-combined latent variables. A segment in the $\x^C$ block is generated conditioned on the latent segment variable of a segment in the block $\x^A$ of the same column, and conditioned on the latent sequence variable of a red-box highlighted segment in the block $\x^B$ of the same row.}
	\label{fig:recomb}
  \end{figure}
  
  Given two segments, $\x^A$ and $\x^B$, we sample a segment $\x^C \sim p(\x | \tilde{\z}_1^A, \tilde{\z}_2^B)$, where $\tilde{\z}_1^A$ is a sampled latent segment variable conditioned on $\x^A$, and $\tilde{\z}_2^B$ is a sampled latent sequence variable conditioned on $\x^B$.
  With a successfully trained FHVAE, $\x^C$ should exhibit the segment-level attributes of $\x^A$, and the sequence-level attributes of $\x^B$.
  Due to space limitations, we only show results of the model trained on the AMI corpus in Figure~\ref{fig:recomb}.
  Eight segments are sampled for $\x^A$, as shown in the upper right corner of the figure.
  Among these segments, the four leftmost ones are close-talking while the rest are far-field, and the first, second, fifth, and sixth from left are female speakers while the rest are males.
  For $\x^B$, three segments of different sequence-level generating factors are sampled, as shown on the left half of the figure. 
  The segments used to infer $\tilde{\z}_2^B$ are highlighted in red boxes; we show the surrounding frames of those segments to better illustrate how sequence-level generating factors affect realization of observations.
  
  Samples of $\x^C$ generated by re-combining latent variables are shown in the lower right corner of the figure.
  It can be clearly observed that $\x^C$ presents the same sequence-level generating factors as $\x^B$,\footnote{In these images, harmonic spacing is the clearest cue for fundamental frequency differences.  Far-field recordings tend to have lower signal-to-noise ratios, which results in blurrier images.} 
  whose latent sequence variable $\x^C$ conditions on.
  Meanwhile, the phonetic content of $\x^C$ stays consistent with $\x^A$,\footnote{Phonetic content can usually be determined by the spectral envelope, and relative position of formants.} whose latent segment variable $\x^C$ conditions on.
  The clear differentiation of generating factors encoded in each sets of latent variables again corroborates the success of our proposed algorithm in training FHVAE models.

  \section{Conclusions and Future Work}
  \label{sec:conclusion}
  In this paper, we discuss the scalability limitations of the original FHVAE training algorithm in terms of runtime, memory, and hyperparameter optimization, and propose a hierarchical sampling algorithm to address this problem.
  Comprehensive study on the memory and time complexity, as well as disentanglement performance verify the effectiveness of the proposed algorithm on all scales of datasets, ranging from 3 to 1,000 hours.
  In the future, we plan to extend FHVAE applications, such as ASR domain adaptation~\cite{hsu2017unsuperviseddomain} and  audio conversion~\cite{hsu2017unsupervised} to larger scales.

  \newpage
  \eightpt
  \bibliographystyle{IEEEtran}

  \bibliography{main.bib}

\end{document}